\documentclass{vldb}
\usepackage{graphicx}
\usepackage{balance}  %
\usepackage{xspace,color}
\usepackage{lettrine}
\usepackage{amsmath,amsfonts}       %

\usepackage{algorithm}
\usepackage{algpseudocode}

\def \ie {\emph{i.e.}\xspace}
\def \eg {\emph{e.g.}\xspace}

\definecolor{darkgreen}{RGB}{0, 140, 0}

\renewcommand{\paragraph}[1]{\medskip \noindent \textbf{#1}}

\usepackage{url}

\begin{document}

\title{Billion-scale similarity search with GPUs}

\numberofauthors{3} %

\author{
\alignauthor
Jeff Johnson\\
 \affaddr{Facebook AI Research}\\
       \affaddr{New York}\\
\alignauthor
Matthijs Douze\\
       \affaddr{Facebook AI Research}\\
       \affaddr{Paris}\\
\alignauthor Herv{\'e} J{\'e}gou\\
       \affaddr{Facebook AI Research}\\
       \affaddr{Paris}\\
}

\maketitle

\begin{abstract}
Similarity search finds application in specialized database systems handling complex data such as images or videos, which are typically represented by high-dimensional features and require specific indexing structures.
This paper tackles the problem of better utilizing GPUs for this task. While GPUs excel at data-parallel tasks, prior approaches are bottlenecked by algorithms that expose less parallelism,
such as $k$-min selection, or make poor use of the memory hierarchy.

We propose a design for $k$-selection that operates at up to 55\% of theoretical peak performance, enabling a nearest neighbor implementation that is 8.5$\times$ faster than prior GPU state of the art.
We apply it in different similarity search scenarios, by proposing optimized design for brute-force, approximate and compressed-domain search based on product quantization. In all these setups, we outperform the state of the art by large margins.
Our implementation enables the construction of a high accuracy $k$-NN graph on 95~million images from the \textsc{Yfcc100M} dataset in 35 minutes, and of a graph connecting 1~billion vectors in less than 12 hours on 4 Maxwell Titan X GPUs.
We have open-sourced our approach\footnote{\texttt{https://github.com/facebookresearch/faiss}} for the sake of comparison and reproducibility.
\end{abstract}

\section{Introduction}

Images and videos constitute a new massive source of data for indexing and search. Extensive metadata for this content is often not available. Search and interpretation of this and other human-generated content, like text, is difficult and important. A variety of machine learning and deep learning algorithms are being used to interpret and classify these complex, real-world entities.
Popular examples include the text representation known as word2vec~\cite{MSCCD13}, representations of images by convolutional neural networks~\cite{RASC14,GWGL14}, and image descriptors for instance search~\cite{GARL16}. Such representations or \textit{embeddings} are usually real-valued, high-dimensional vectors of 50 to 1000+ dimensions.
Many of these vector representations can only effectively be produced on GPU systems, as the underlying processes either have high arithmetic complexity and/or high data bandwidth demands~\cite{Krizhevsky12}, or cannot be effectively partitioned without failing due to communication overhead or representation quality~\cite{Shamir14}. Once produced, their manipulation is itself arithmetically intensive.
However, how to utilize GPU assets is not straightforward. More generally, how to exploit new heterogeneous architectures is a key subject for the database community~\cite{boncz2016special}.

In this context, searching by numerical \textit{similarity} rather than via structured relations is more suitable. This could be to find the most similar content to a picture, or to find the vectors that have the highest response to a linear classifier on all vectors of a collection.

One of the most expensive operations to be performed on large collections is to compute a $k$-NN graph. It is a directed graph where each vector of the database is a node and each edge connects a node to its $k$ nearest neighbors. This is our flagship application. Note, state of the art methods like NN-Descent~\cite{DCL11} have a large memory overhead on top of the dataset itself and cannot readily scale to the billion-sized databases we consider.

Such applications must deal with the \textit{curse of dimensionality}~\cite{WSB98}, rendering both exhaustive search or exact indexing for non-exhaustive search impractical on billion-scale databases. This is why there is a large body of work on approximate search and/or graph construction.
To handle huge datasets that do not fit in RAM, several approaches employ an internal compressed representation of the vectors using an encoding. This is especially convenient for memory-limited devices like GPUs. It turns out that accepting a minimal accuracy loss results in orders of magnitude of compression~\cite{HMD15}. The most popular vector compression methods can be classified into either binary codes~\cite{GL11,HWS13}, or quantization methods~\cite{JDS11,PJA10}. Both have the desirable property that searching neighbors does not require reconstructing the vectors.

Our paper focuses on methods based on product quantization (PQ) codes, as these were shown to be more effective than binary codes~\cite{NF13}. In addition, binary codes incur important overheads for non-exhaustive search methods~\cite{NPF12}.
Several improvements were proposed after the original product quantization proposal known as IVFADC~\cite{JDS11}; most are difficult to implement efficiently on GPU.
For instance, the inverted multi-index~\cite{BL12}, useful for high-speed/low-quality operating points, depends on a complicated ``multi-sequence'' algorithm. The optimized product quantization or OPQ~\cite{GHKS14} is a linear transformation on the input vectors that improves the accuracy of the product quantization; it can be applied as a pre-processing. The SIMD-optimized IVFADC implementation from~\cite{andre15} operates only with sub-optimal parameters (few coarse quantization centroids). %
Many other methods, like LOPQ and the Polysemous codes~\cite{KA14,DJP16} are too complex to be implemented efficiently on GPUs.

There are many implementations of similarity search on GPUs, but mostly with binary codes~\cite{Pan11}, small datasets~\cite{Wakatani14}, or exhaustive search~\cite{Dashti13, Sismanis12, Tang15}. To the best of our knowledge, only the work by Wieschollek et al.~\cite{WieschollekCVPR16} appears suitable for billion-scale datasets with quantization codes. This is the prior state of the art on GPUs, which we compare against in Section~\ref{sec:largescale}.
\medskip

This paper makes the following contributions:

\begin{itemize}
\item a GPU $k$-selection algorithm, operating in fast register memory and flexible enough to be fusable with other kernels, for which we provide a complexity analysis;
\item a near-optimal algorithmic layout for exact and approximate $k$-nearest neighbor search on GPU;
\item a range of experiments that show that these improvements outperform previous art by a large margin on mid- to large-scale nearest-neighbor search tasks, in single or multi-GPU configurations.
\end{itemize}

The paper is organized as follows. Section~\ref{sec:problem} introduces the context and notation. Section~\ref{sec:GPU} reviews GPU architecture and discusses problems appearing when using it for similarity search.
Section~\ref{sec:ourGPUkselect} introduces one of our main contributions, \ie, our k-selection method for GPUs, while Section~\ref{sec:layout} provides details regarding the algorithm computation layout.
Finally, Section~\ref{sec:experiments} provides extensive experiments for our approach, compares it to the state of the art, and shows concrete use cases for image collections.

\section{Problem statement}
\label{sec:problem}

\newcommand{\nq}{ {n_{\mathrm{q}} }}

We are concerned with similarity search in vector collections. Given the query vector $x \in \mathbb{R}^d$ and the collection\footnote{To avoid clutter in 0-based indexing, we use the array notation $0:\ell$ to denote the range $\{0,...,\ell-1\}$ inclusive.} $[y_i]_{i=0:\ell}~ (y_i \in \mathbb{R}^{d})$, we search:
\begin{equation}
L = k\text{-}\textrm{argmin}_{i=0:\ell} \|x - y_i\|_2,
\label{eq:dmin}
\end{equation}
\textit{i.e.}, we search the $k$ nearest neighbors of $x$ in terms of L2 distance. The L2 distance is used most often, as it is optimized by design when learning several embeddings (\eg, \cite{GARL16}), due to its attractive linear algebra properties.

The lowest distances are  collected by $k$-selection. For an array $[a_i]_{i=0:\ell}$, $k$-selection finds the $k$ lowest valued elements $[a_{s_i}]_{i=0:k}$, $a_{s_i} \leq a_{s_{i+1}}$, along with the indices $[s_i]_{i=0:k}$, $0 \leq s_i < \ell$, of those elements from the input array. The $a_i$ will be 32-bit floating point values; the $s_i$ are 32- or 64-bit integers.  Other comparators are sometimes desired; \eg, for cosine similarity we search for \emph{highest} values. The order between equivalent keys $a_{s_i} = a_{s_j}$ is not specified.

\paragraph{Batching.}
Typically, searches are performed in batches of $\nq$ query vectors $[x_j]_{j=0:\nq}~ (x_j \in \mathbb{R}^{d})$ in parallel, which allows for more flexibility when executing on multiple CPU threads or on GPU.
Batching for $k$-selection entails selecting $\nq \times k$ elements and indices from $\nq$ separate arrays, where each array is of a potentially different length $\ell_i \geq k$.

\newcommand{\one}{\mathbf{1}}

\paragraph{Exact search.}
The exact solution computes the full pairwise distance matrix $D = [\|x_j - y_i\|_2^2]_{j=0:\nq,i=0:\ell} \in \mathbb{R}^{\nq\times\ell}$. In practice, we use the decomposition

\begin{equation}
\|x_j - y_i\|_2^2 = \|x_j\|^2 + \|y_i\|^2 - 2 \langle x_j, y_i \rangle.
\label{eq:dminmmul}
\end{equation}
The two first terms can be precomputed in one pass over the matrices $X$ and $Y$ whose rows are the $[x_j]$ and $[y_i]$. The bottleneck is to evaluate $\langle x_j, y_i \rangle$, equivalent to the matrix multiplication $X Y^\top$. The $k$-nearest neighbors for each of the $n_q$ queries are $k$-selected along each row of $D$.

\paragraph{Compressed-domain search.}
From now on, we focus on approximate nearest-neighbor search. We consider, in particular, the \textit{IVFADC} indexing structure~\cite{JDS11}.
The IVFADC index relies on two levels of quantization, and the database vectors are encoded. The database vector $y$ is approximated as:
\begin{equation}
y \approx q(y) = q_1(y) + q_2 (y - q_1(y))
\label{eq:ivfpqquant}
\end{equation}
where $q_1 : \mathbb{R}^d \rightarrow \mathcal{C}_1 \subset \mathbb{R}^d$ and $q_2 : \mathbb{R}^d \rightarrow \mathcal{C}_2\subset \mathbb{R}^d$ are quantizers; \ie, functions that output an element from a finite set. Since the sets are finite, $q(y)$ is encoded as the index of $q_1(y)$ and that of $q_2(y - q_1(y))$. The first-level quantizer is a \textit{coarse quantizer} and the second level \textit{fine quantizer} encodes the residual vector after the first level.

The Asymmetric Distance Computation (ADC) search method returns an approximate result:
\begin{equation}
L_\mathrm{ADC} = k\text{-}\textrm{argmin}_{i=0:\ell} \|x - q(y_i)\|_2.
\end{equation}

For IVFADC the search is not exhaustive. Vectors for which the distance is computed are pre-selected depending on the first-level quantizer $q_1$:
\newcommand{\nprobe}{\tau}
\begin{equation}
L_\mathrm{IVF} = \nprobe\text{-}\textrm{argmin}_{c \in \mathcal{C}_1} \| x - c \|_2.
\label{eq:livf}
\end{equation}
The \textit{multi-probe parameter} $\tau$ is the number of coarse-level centroids we consider. The quantizer operates a nearest-neighbor search with exact distances, in the set of reproduction values. %
Then, the IVFADC search computes

\begin{equation}
L_\mathrm{IVFADC} = \underset{i=0:\ell \textrm{ s.t. } q_1(y_i) \in L_\mathrm{IVF}}{k\text{-}\textrm{argmin}} \|x - q(y_i)\|_2.
\label{eq:ivfpqargmin}
\end{equation}
Hence, IVFADC  relies on the same distance estimations as the two-step quantization of ADC, but computes them only on a subset of vectors.

\newcommand{\invlist}{\mathcal{I}}

The corresponding data structure, the \emph{inverted file}, groups the vectors $y_i$ into $|\mathcal{C}_1|$ \emph{inverted lists} $\invlist_1, ..., \invlist_{|\mathcal{C}_1|}$ with homogeneous $q_1(y_i)$. Therefore, the most memory-intensive operation is computing $L_\mathrm{IVFADC}$, and boils down to linearly scanning $\nprobe$ inverted lists.

\paragraph{The quantizers.}
The quantizers $q_1$ and $q_2$ have different properties. $q_1$ needs to have a relatively low number of reproduction values so that the number of inverted lists does not explode. We typically use $|C_1| \approx \sqrt{\ell}$, trained via $k$-means. For $q_2$, we can afford to spend more memory for a more extensive representation. The ID of the vector (a 4- or 8-byte integer) is also stored in the inverted lists, so it makes no sense to have shorter codes than that; \textit{i.e.}, $\log_2 |\mathcal{C}_2| > 4\times 8$.

\paragraph{Product quantizer.}
We use a \textit{product quantizer}~\cite{JDS11} for $q_2$, which provides a large number of reproduction values without increasing the processing cost.
It interprets the vector $y$ as $b$ sub-vectors $y=[y^0 ... y^{b-1}]$, where $b$ is an even divisor of the dimension $d$. Each sub-vector is quantized with its own quantizer, yielding the tuple $(q^0(y^0),$ ..., $q^{b-1}(y^{b-1}))$. The sub-quantizers typically have 256 reproduction values, to fit in one byte. The quantization value of the product quantizer is then $q_2(y) = q^0(y^0) + 256 \times q^1 (y^1) + ... + 256^{b-1}\times q^{b-1}$, which from a storage point of view is just the concatenation of the bytes produced by each sub-quantizer. Thus,  the product quantizer generates $b$-byte codes with $|\mathcal{C}_2| = 256^b$ reproduction values. The $k$-means dictionaries of the quantizers are small and quantization is computationally cheap.

\section{GPU: overview and k-selection}
\label{sec:GPU}

This section reviews salient details of Nvidia's general-purpose GPU architecture and programming model~\cite{Lindholm08}. We then focus on one of the less GPU-compliant parts involved in similarity search, namely the $k$-selection, and discuss the literature and challenges.

\subsection{Architecture}

\paragraph{GPU lanes and warps.} The Nvidia GPU is a general-purpose computer that executes instruction streams using a 32-wide vector of \textit{CUDA threads} (the \textit{warp}); individual threads in the warp are referred to as \textit{lanes}, with a \emph{lane ID} from 0\,--\,31. Despite the ``thread'' terminology, the best analogy to modern vectorized multicore CPUs is that each warp is a separate CPU hardware thread, as the warp shares an instruction counter. Warp lanes taking different execution paths results in \textit{warp divergence}, reducing performance. Each lane has up to 255 32-bit registers in a shared register file. The CPU analogy is that there are up to 255 vector registers of width 32, with warp lanes as SIMD vector lanes.

\paragraph{Collections of warps.} A user-configurable collection of 1 to 32 warps comprises a \textit{block} or a \textit{co-operative thread array} (CTA). Each block has a high speed \textit{shared memory}, up to 48\,KiB in size. Individual CUDA threads have a block-relative ID, called a \textit{thread id}, which can be used to partition and assign work. Each block is run on a single core of the GPU called a \textit{streaming multiprocessor} (SM).
Each SM has \textit{functional units}, including ALUs, memory load/store units, and various special instruction units. %
A GPU hides execution latencies by having many operations in flight on warps across all SMs. Each individual warp lane instruction throughput is low and latency is high, but the aggregate arithmetic throughput of all SMs together %
is 5\,--\,10$\times$ higher than typical CPUs.

\paragraph{Grids and kernels.} Blocks are organized in a \textit{grid} of blocks in a \textit{kernel}. Each block is assigned a grid relative ID. The kernel is the unit of work (instruction stream with arguments) scheduled by the host CPU for the GPU to execute. %
After a block runs through to completion, new blocks can be scheduled. Blocks from different kernels can run concurrently. Ordering between kernels is controllable via ordering primitives such as \textit{streams} and \textit{events}.

\paragraph{Resources and occupancy.} The number of blocks executing concurrently depends upon shared memory and register resources used by each block. Per-CUDA thread register usage is determined at compilation time, while shared memory usage can be chosen at runtime. This usage affects \textit{occupancy} on the GPU. %
If a block demands all 48\,KiB of shared memory for its private usage, or 128 registers per thread as opposed to 32, then only 1\,--\,2 other blocks can run concurrently on the same SM, resulting in low occupancy. %
Under high occupancy more blocks will be present across all SMs, allowing more work to be in flight at once.

\paragraph{Memory types.} Different blocks and kernels communicate through \textit{global memory}, typically 4\,--\,32 GB in size, with 5\,--\,10$\times$ higher bandwidth than CPU main memory. %
Shared memory is analogous to CPU L1 cache in terms of speed. GPU register file memory is the highest bandwidth memory.
In order to maintain the high number of instructions in flight on a GPU, a vast register file is also required: 14\,MB in the latest Pascal P100, in contrast with a few tens of \,KB on CPU. %
A ratio of 250\,:\,6.25\,:\,1 for register to shared to global memory aggregate cross-sectional bandwidth is typical on GPU, yielding 10\,--\,100s of TB/s for the register file~\cite{Canny13}.

\subsection{GPU register file usage}

\paragraph{Structured register data.}
Shared and register memory usage involves efficiency tradeoffs; they lower occupancy but can increase overall performance by retaining a larger working set in a faster memory. Making heavy use of register-resident data at the expense of occupancy or instead of shared memory is often profitable~\cite{Volkov08}.

As the GPU register file is very large, storing structured data (not just temporary operands) is useful. A single lane can use its (scalar) registers to solve a local task, but with limited parallelism and storage. %
Instead, lanes in a GPU warp can instead exchange register data using the \textit{warp shuffle} instruction, enabling warp-wide parallelism and storage.

\paragraph{Lane-stride register array.}
A common pattern to achieve this is a \textit{lane-stride register array}. That is, given elements $[a_i]_{i = 0:\ell}$, each successive value is held in a register by neighboring lanes. The array is stored in $\ell / 32$ registers per lane, with $\ell$ a multiple of 32. Lane $j$ stores $\{ a_j, a_{32 + j}, ..., a_{\ell-32 + j}\}$, while register $r$ holds $\{a_{32r}, a_{32r + 1}, ..., a_{32r + 31}\}$. %

For manipulating the $[a_i]$, the register in which $a_i$ is stored (\textit{i.e.,} $\left \lfloor i / 32 \right \rfloor$) and $\ell$ must be known at assembly time, while the lane (\textit{i.e.,} $i \bmod 32$) can be runtime knowledge. A wide variety of access patterns (shift, any-to-any) are provided; we use the butterfly permutation~\cite{Leighton92} extensively.

\subsection{k-selection on CPU versus GPU}

$k$-selection algorithms, often for arbitrarily large $\ell$ and $k$, can be translated to a GPU, including \textit{radix selection} and \textit{bucket selection}~\cite{Alabi12}, \textit{probabilistic selection}~\cite{Monroe11}, \textit{quickselect}~\cite{Dashti13}, and \textit{truncated sorts}~\cite{Sismanis12}. Their performance is dominated by multiple passes over the input in global memory. Sometimes for similarity search, the input distances are computed on-the-fly or stored only in small blocks, not in their entirety. The full, explicit array might be too large to fit into any memory, and its size could be unknown at the start of the processing, rendering algorithms that require multiple passes impractical.
They suffer from other issues as well. Quickselect requires partitioning on a storage of size $\mathcal O(\ell)$, a data-dependent memory movement. This can result in excessive memory transactions, or requiring parallel prefix sums to determine write offsets, with synchronization overhead. Radix selection has no partitioning but multiple passes are still required.

\paragraph{Heap parallelism.} In similarity search applications, one is usually interested only in a small number of results, $k<1000$ or so.
In this regime, selection via max-heap is a typical choice on the CPU, but heaps do not expose much data parallelism (due to serial tree update) and cannot saturate SIMD execution units. The \textit{ad-heap}~\cite{Liu14} takes better advantage of parallelism available in heterogeneous systems, but still attempts to partition serial and parallel work between appropriate execution units. Despite the serial nature of heap update, for small $k$ the CPU can maintain all of its state in the L1 cache with little effort, and L1 cache latency and bandwidth remains a limiting factor. Other similarity search components, like PQ code manipulation, tend to have greater impact on CPU performance~\cite{andre15}.

\paragraph{GPU heaps.} Heaps can be similarly implemented on a GPU~\cite{Barrientos11}. However, a straightforward GPU heap implementation  suffers from high warp divergence and irregular, data-dependent memory movement, since the path taken for each inserted element depends upon other values
in the heap. %

GPU parallel priority queues \cite{He2012} improve over the serial heap update by allowing multiple concurrent updates, but they require a potential number of small sorts for each insert and data-dependent memory movement. Moreover, it uses multiple synchronization barriers through kernel launches in different streams, plus the additional latency of successive kernel launches and coordination with the CPU host.

Other more novel GPU algorithms are available for small $k$, namely the selection algorithm in the \textit{fgknn} library~\cite{Tang15}. This is a complex algorithm that may suffer from too many synchronization points, greater kernel launch overhead, usage of slower memories, excessive use of hierarchy, partitioning and buffering. However, we take inspiration from this particular algorithm through the use of parallel merges as seen in their \emph{merge queue} structure.

\section{Fast k-selection on the GPU}
\label{sec:ourGPUkselect}

\newcommand{\warpbased}{\textsc{WarpSelect}\xspace}

For any CPU or GPU algorithm, either memory or arithmetic throughput should be the limiting factor as per the \textit{roofline performance model}~\cite{Williams09}. For input from global memory, $k$-selection cannot run faster than the time required to scan the input once at peak memory bandwidth. We aim to get as close to this limit as possible. Thus, we wish to perform a single pass over the input data (from global memory or produced on-the-fly, perhaps fused with a kernel that is generating the data).

We want to keep intermediate state in the fastest memory: the register file.
The major disadvantage of register memory is that the indexing into the register file must be known at assembly time, which is a strong constraint on the algorithm. %

\subsection{In-register sorting}

 We use an in-register sorting primitive as a building block. Sorting networks are commonly used on SIMD architectures~\cite{Chhugani08}, as they exploit vector parallelism. They are easily implemented on the GPU, and we build sorting networks with lane-stride register arrays.

We use a variant of \textit{Batcher's bitonic sorting network}~\cite{Batcher68}, which is a set of parallel merges on an array of size~$2^k$. Each merge takes $s$ arrays of length $t$ ($s$ and $t$ a power of 2) to $s/2$ arrays of length $2t$, using $\log_2(t)$ parallel steps. A bitonic sort applies this merge recursively: to sort an array of length $\ell$, merge $\ell$ arrays of length $1$ to $\ell/2$ arrays of length $2$, to $\ell/4$ arrays of length $4$, successively to $1$ sorted array of length $\ell$, leading to $\frac{1}{2}(\log_2(\ell)^2 + \log_2(\ell))$ parallel merge steps.

\begin{algorithm}[t]
\algblockdefx{parallelfor}{endparallelfor}[1]
  {\textbf{parallel for }#1 \textbf{do}}
  {\textbf{end for}}

\algblockdefx{paralleldo}{endparalleldo}
  {\textbf{parallel do}}
  {\textbf{end do}}

  \caption{Odd-size merging network}\label{alg:odd-merge}
  \begin{algorithmic}
    \Function{merge-odd}{$[L_i]_{i=0:\ell_L}, [R_i]_{i=0:\ell_R}$}
    \parallelfor{$i \gets 0: \min(\ell_L, \ell_R)$}
    \State \Comment inverted 1st stage; inputs are already sorted
    \State \Call{compare-swap}{$L_{\ell_L - i - 1}, R_{i}$}
    \endparallelfor
    \paralleldo
    \State \Comment If $\ell_L = \ell_R$ and a power-of-2, these are equivalent
    \State \Call{merge-odd-continue}{$[L_i]_{i=0:\ell_L}$, \texttt{left}}
    \State \Call{merge-odd-continue}{$[R_i]_{i=0:\ell_R}$, \texttt{right}}
    \endparalleldo
    \EndFunction

    \Function{merge-odd-continue}{$[x_i]_{i=0:\ell}, p$}
    \If{$\ell > 1$}
    \State {$h \gets 2^{\left \lceil \log_2 \ell \right \rceil - 1}$} \Comment largest power-of-2 $< \ell$
    \parallelfor{$i \gets 0: \ell - h$}
    \State \Comment Implemented with warp shuffle butterfly
    \State \Call{compare-swap}{$x_i, x_{i + h}$}
    \endparallelfor
    \paralleldo
    \If{$p$ = \texttt{left}} \Comment left side recursion
    \State \Call{merge-odd-continue}{$[x_i]_{i=0:\ell - h}, \texttt{left}$}
    \State \Call{merge-odd-continue}{$[x_i]_{i=\ell - h:\ell}, \texttt{right}$}
    \Else \Comment right side recursion
    \State \Call{merge-odd-continue}{$[x_i]_{i=0:h}, \texttt{left}$}
    \State \Call{merge-odd-continue}{$[x_i]_{i=h:\ell}, \texttt{right}$}
    \EndIf
    \endparalleldo
    \EndIf
    \EndFunction

  \end{algorithmic}
\end{algorithm}

\begin{figure}
\begin{center}
\includegraphics[width=0.9\linewidth]{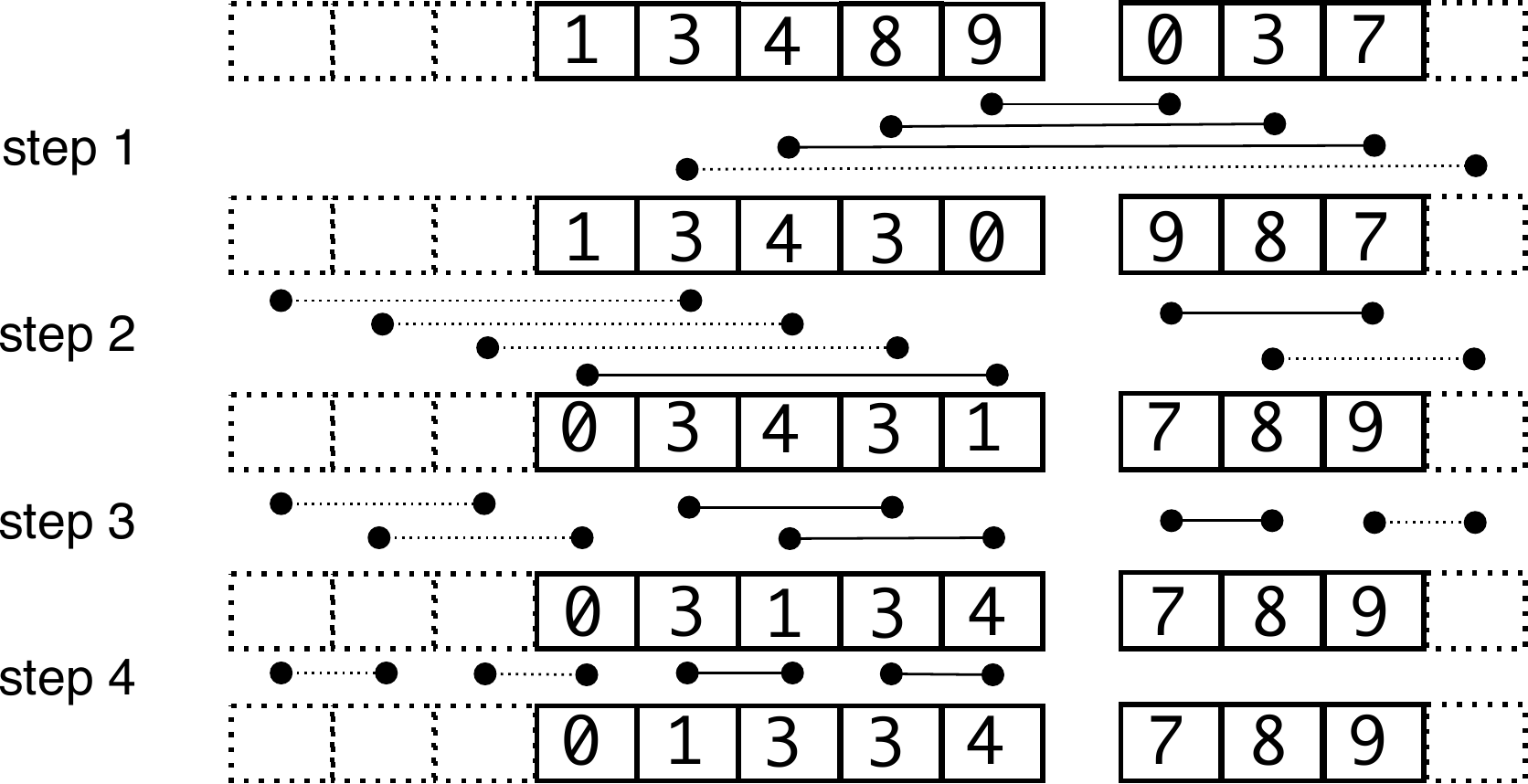}
\end{center}
\caption{\label{fig:odd-merge}
	Odd-size network merging arrays of sizes 5 and 3.
        Bullets indicate parallel compare/swap. Dashed lines are elided elements or comparisons.
}
\end{figure}

\paragraph{Odd-size merging and sorting networks.}
If some input data is already sorted, we can modify the network to avoid merging steps. We may  also not have a full power-of-2 set of data, in which case we can efficiently shortcut to deal with the smaller size.

Algorithm~\ref{alg:odd-merge} is an odd-sized merging network that merges already sorted \textit{left} and \textit{right} arrays, each of arbitrary length. While the bitonic network merges \textit{bitonic} sequences, we start with \textit{monotonic} sequences: sequences sorted monotonically. A bitonic merge is made monotonic by reversing the first comparator stage.

The odd size algorithm is derived by considering arrays to be padded to the next highest power-of-2 size with dummy elements that are never swapped (the merge is monotonic) and are already properly positioned; any comparisons with dummy elements are elided. A left array is considered to be padded with dummy elements at the start; a right array has them at the end. A merge of two sorted arrays of length $\ell_L$ and $\ell_R$ to a sorted array of $\ell_L + \ell_R$ requires $\left \lceil \log_2(\max(\ell_L, \ell_R)) \right \rceil + 1$ parallel steps. Figure~\ref{fig:odd-merge} shows Algorithm~\ref{alg:odd-merge}'s merging network for arrays of size 5 and 3, with 4 parallel steps.

The \textsc{compare-swap} is implemented using warp shuffles on a lane-stride register array. Swaps with a stride a multiple of 32 occur directly within a lane as the lane holds both elements locally. Swaps of stride $\leq 16$ or a non-multiple of 32 occur with warp shuffles. In practice, used array lengths are multiples of 32 as they are held in lane-stride arrays.

\begin{algorithm}

\algblockdefx{parallelfor}{endparallelfor}[1]
  {\textbf{parallel for }#1 \textbf{do}}
  {\textbf{end for}}

\algblockdefx{paralleldo}{endparalleldo}
  {\textbf{parallel do}}
  {\textbf{end do}}

  \caption{Odd-size sorting network}\label{alg:odd-sort}
  \begin{algorithmic}
    \Function{sort-odd}{$[x_i]_{i=0:\ell}$}
    \If{$\ell > 1$}
    \paralleldo
    \State \Call{sort-odd}{$[x_i]_{i=0:\left \lfloor \ell / 2 \right \rfloor}$}
    \State \Call{sort-odd}{$[x_i]_{i={\left \lfloor \ell / 2 \right \rfloor}:\ell}$}
    \endparalleldo
    \State \Call{merge-odd}{$[x_i]_{i=0:\left \lfloor \ell / 2 \right \rfloor}, [x_i]_{i={\left \lfloor \ell / 2 \right \rfloor}:\ell}$}
    \EndIf
    \EndFunction
  \end{algorithmic}
\end{algorithm}

Algorithm~\ref{alg:odd-sort} extends the merge to a full sort. Assuming no structure present in the input data, $\frac{1}{2}(\left \lceil \log_2(\ell) \right \rceil^2 + \left \lceil \log_2(\ell) \right \rceil)$ parallel steps are required for sorting data of length $\ell$.

\subsection{WarpSelect}

Our $k$-selection implementation, \warpbased, maintains state entirely in registers, requires only a single pass over data and avoids cross-warp synchronization. It uses \textsc{merge-odd} and \textsc{sort-odd} as primitives. Since the register file provides much more storage than shared memory, it supports $k \leq 1024$. %
Each warp is dedicated to $k$-selection to a single one of the $n$ arrays $[a_i]$. If $n$ is large enough, a single warp per each $[a_i]$ will result in full GPU occupancy. Large $\ell$ per warp is handled by recursive decomposition, if $\ell$ is known in advance.

\newcommand{\RTK}{T}
\newcommand{\RTV}{V^\mathrm{T}}
\newcommand{\RTKp}{K^\mathrm{T}}
\newcommand{\RTVp}{V^\mathrm{T}}

\newcommand{\RWV}{V^\mathrm{W}}
\newcommand{\RWK}{W}

\begin{figure}[t]
\includegraphics[width=\linewidth]{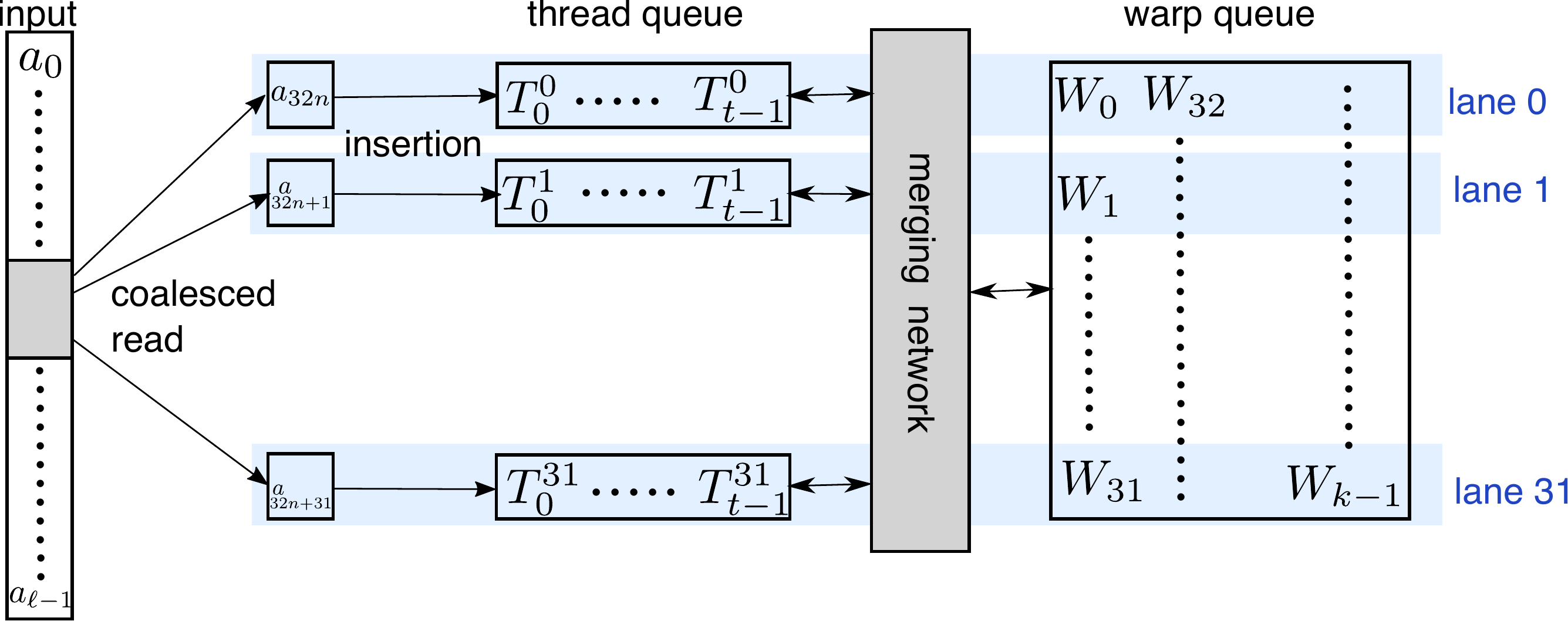}
\caption{\label{fig:algoover}
	Overview of \warpbased. The input values stream in on the left, and the warp queue on the right holds the output result.
}
\end{figure}

\paragraph{Overview.}
Our approach (Algorithm~\ref{alg:warp-select} and Figure~\ref{fig:algoover}) operates on values, with associated indices carried along (omitted from the description for simplicity). It selects the $k$ least values that come from global memory, or from intermediate value registers if fused into another kernel providing the values. Let $[a_i]_{i=0:\ell}$ be the sequence provided for selection.

The elements (on the left of Figure~\ref{fig:algoover}) are processed in groups of 32, the warp size. Lane $j$ is responsible for processing $\{a_j, a_{32+j}, ...\}$; thus, if the elements come from global memory, the reads are contiguous and coalesced into a minimal number of memory transactions. %

\begin{algorithm}[b]
\algblockdefx{parallelfor}{endparallelfor}[1]
  {\paragraph{parallel for }#1 \paragraph{do}}
  {\paragraph{end for}}

\algblockdefx{paralleldo}{endparalleldo}
  {\paragraph{parallel do}}
  {\paragraph{end do}}

  \caption{\warpbased pseudocode for lane $j$}\label{alg:warp-select}
  \begin{algorithmic}
    \Function{WarpSelect}{$a$}
    \If{$a < \RTK_0^j$}
    \State insert $a$ into our $[\RTK_i^j]_{i=0:t}$
    \EndIf
    \If{\textsc{warp-ballot}($\RTK_{0}^{j} < \RWK_{k - 1})$} %
    \State \Comment Reinterpret thread queues as lane-stride array
    \State $[\alpha_i]_{i=0:32t} \gets \textsc{cast}([\RTK_i^j]_{i=0:t, j=0:32})$
   	\State \Comment concatenate and sort thread queues
    \State \Call{sort-odd}{$[\alpha_i]_{i=0:32t}$}
    \State \Call{merge-odd}{$[\RWK_i]_{i=0:k}, [\alpha_i]_{i=0:32t}$}
    \State \Comment Reinterpret lane-stride array as thread queues
    \State $[\RTK_i^j]_{i=0:t,j=0:32} \gets \textsc{cast}([\alpha_i]_{i=0:32t})$
    \State \Call{reverse-array}{$[\RTK_i]_{i=0:t}$}
    \State \Comment Back in thread queue order, invariant restored
    \EndIf
    \EndFunction
  \end{algorithmic}
\end{algorithm}

\paragraph{Data structures.}
Each lane $j$ maintains a small queue of $t$ elements in registers, called the \textit{thread queues} $[\RTK_i^j]_{i=0:t}$, ordered  from largest to smallest ($\RTK_i^j \geq \RTK_{i+1}^j$). The choice of $t$ is made relative to $k$, see Section~\ref{sec:k-sel-complexity}. The thread queue is a first-level filter for new values coming in. If a new $a_{32i+j}$ is greater than the largest key currently in the queue, $\RTK_0^j$, it is guaranteed that it won't be in the $k$ smallest final results.

The warp shares a lane-stride register array of $k$ smallest seen elements, $[\RWK_i]_{i=0:k}$, called the \textit{warp queue}. It is ordered from smallest to largest ($\RWK_i \leq \RWK_{i+1}$); if the requested $k$ is not a multiple of 32, we round it up.
This is a second level data structure that will be used to maintain all of the $k$ smallest warp-wide seen values. The thread and warp queues are initialized to maximum sentinel values, \textit{e.g.},~$+\infty$.

\paragraph{Update.}
The three invariants maintained are:

\begin{itemize}
\item
  all per-lane $\RTK_0^j$ are not in the min-$k$
\item
  all per-lane $\RTK_0^j$ are greater than all warp queue keys $\RWK_i$
\item
  all $a_i$ seen so far in the min-$k$ are contained in either some lane's thread queue ($[\RTK_i^j]_{i=0:t, j=0:32}$), or in the warp queue.
\end{itemize}

Lane $j$ receives a new $a_{32i+j}$ and attempts to insert it into its thread queue. If $a_{32i+j} > \RTK_0^j$, then the new pair is by definition not in the $k$ minimum, and can be rejected.

Otherwise, it is inserted into its proper sorted position in the thread queue, thus ejecting the old $\RTK_0^j$. All lanes complete doing this with their new received pair and their thread queue, but it is now possible that the second invariant have been violated. Using the \textit{warp ballot} instruction, we determine if any lane has violated the second invariant. If not, we are free to continue processing new elements.

\paragraph{Restoring the invariants.}
If any lane has its invariant violated, then the warp uses \textsc{odd-merge} to merge and sort the thread and warp queues together. The new warp queue will be the min-$k$ elements across the merged, sorted queues, and the new thread queues will be the remainder, from min-$(k+1)$ to min-$(k + 32 t + 1)$. This restores the invariants and we are free to continue processing subsequent elements.

Since the thread and warp queues are already sorted, we merge the sorted warp queue of length $k$ with 32 sorted arrays of length $t$. Supporting odd-sized merges is important because Batcher's formulation would require that $32t = k$ and is a power-of-2; thus if $k = 1024$, $t$ must be 32. We found that the optimal $t$ is way smaller (see below).

Using \textsc{odd-merge} to merge the 32 already sorted thread queues would require a \textit{struct-of-arrays} to \textit{array-of-structs} transposition in registers across the warp, since the $t$ successive sorted values are held in different registers in the same lane rather than a lane-stride array. This is possible~\cite{Catanzaro14}, but would use a comparable number of warp shuffles, so we just reinterpret the thread queue registers as an (unsorted) lane-stride array and sort from scratch. Significant speedup is realizable by using \textsc{odd-merge} for the merge of the aggregate sorted thread queues with the warp queue.

\paragraph{Handling the remainder.}
If there are remainder elements because $\ell$ is not a multiple of 32, those are inserted into the thread queues for the lanes that have them, after which we proceed to the output stage.

\paragraph{Output.}
A final sort and merge is made of the thread and warp queues, after which the warp queue holds all min-$k$ values.

\subsection{Complexity and parameter selection}
\label{sec:k-sel-complexity}

For each incoming group of 32 elements, \warpbased can perform 1, 2 or 3 constant-time operations, all happening in warp-wide parallel time:
\begin{enumerate}
\item
	read 32 elements, compare to all thread queue heads $T_0^j$, cost $C_1$, happens $N_1$ times;
\item
 	if $\exists j\in \{0, ..., 31\}$, $a_{32n+j} < T_0^j$, perform insertion sort on those specific thread queues, cost $C_2=\mathcal{O}(t)$, happens $N_2$ times;
\item
  if $\exists j, T_0^j < W_{k-1}$, sort and merge queues, cost $C_3 = \mathcal{O}(t\log(32t)^2 + k\log(\max(k, 32t)))$, happens $N_3$ times.
\end{enumerate}

Thus, the total cost is $N_1 C_1 + N_2 C_2 + N_3 C_3$. $N_1 = \ell / 32$, and on random data drawn independently, $N_2 =\mathcal{O}(k \log(\ell))$ and $N_3 = \mathcal{O}(k \log(\ell) / t)$, see the Appendix for a full derivation.
Hence, the trade-off is to balance a cost in $N_2C_2$ and one in $N_3C_3$.
The practical choice for $t$ given $k$ and $\ell$ was made by experiment on a variety of $k$-NN data. For $k \leq 32$, we use $t = 2$, $k \leq 128$ uses $t = 3$, $k \leq 256$ uses $t = 4$, and $k \leq 1024$ uses $t = 8$, all irrespective of $\ell$.

\section{Computation layout}
\label{sec:layout}

This section explains how IVFADC, one of the indexing methods originally built upon product quantization~\cite{JDS11}, is implemented efficiently. Details on distance computations and articulation with $k$-selection are the key to understanding why this method can outperform more recent GPU-compliant approximate nearest neighbor strategies~\cite{WieschollekCVPR16}.

\subsection{Exact search}
\label{sec:exactimpl}

We briefly come back to the exhaustive search method, often referred to as exact brute-force. It is interesting on its own for exact nearest neighbor search in small datasets. It is also a component of many indexes in the literature. In our case, we use it for the IVFADC coarse quantizer $q_1$.

As stated in Section~\ref{sec:problem}, the distance computation boils down to a matrix multiplication. We use optimized GEMM routines in the cuBLAS library to calculate the $ - 2 \langle x_j, y_i \rangle$ term for L2 distance, resulting in a partial distance matrix $D'$. To complete the distance calculation, we use a fused $k$-selection kernel that adds the $\|y_i\|^2$ term to each entry of the distance matrix and immediately submits the value to $k$-selection in registers. The $\|x_j\|^2$ term need not be taken into account before $k$-selection. Kernel fusion thus allows for only 2 passes (GEMM write, $k$-select read) over $D'$, compared to other implementations that may require 3 or more.
Row-wise $k$-selection is likely not fusable with a well-tuned GEMM kernel, or would result in lower overall efficiency.

As $D'$ does not fit in GPU memory for realistic problem sizes, the problem is tiled over the batch of queries, with $t_q \leq n_q$ queries being run in a single tile. Each of the $\left \lceil n_q / t_q \right \rceil$ tiles are independent problems, but we run two in parallel on different streams to better occupy the GPU, so the effective memory requirement of $D$ is $\mathcal{O}(2 \ell t_q)$. The computation can similarly be tiled over $\ell$. For very large input coming from the CPU, we support buffering with pinned memory to overlap CPU to GPU copy with GPU compute.

\subsection{IVFADC indexing}

\paragraph{PQ lookup tables.}
At its core, the IVFADC requires computing the distance from a vector to a set of product quantization reproduction values. By developing Equation~(\ref{eq:ivfpqargmin}) for a database vector $y$, we obtain:
\begin{equation}
\|x- q(y)\|^2_2 = \|x - q_1(y) - q_2(y-q_1(y))\|^2_2.
\label{eq:twostagedecomp}
\end{equation}
If we decompose the residual vectors left after $q_1$ as:
\begin{eqnarray}
y - q_1(y) &=& [
\widetilde{y^1} \cdots
\widetilde{y^b}]
\textrm{ and }\\
x - q_1(y) &=& [
\widetilde{x^1} \cdots
\widetilde{x^b}]
\end{eqnarray}
then the distance is rewritten as:
\begin{equation}
\|x- q(y)\|^2_2 =
\|\widetilde{x^1} - q^1(\widetilde{y^1})\|_2^2 + ... +
\|\widetilde{x^b} - q^b(\widetilde{y^b})\|_2^2.
\label{eq:pqsum}
\end{equation}
Each quantizer $q^1, ..., q^b$ has 256 reproduction values, so when $x$ and $q_1(y)$ are known all distances can be precomputed and stored in tables $T_1,..., T_b$ each of size 256~\cite{JDS11}.
Computing the sum~(\ref{eq:pqsum}) consists of $b$ look-ups and additions. Comparing the cost to compute $n$ distances:
\begin{itemize}
\item
	Explicit computation: $n\times d$ mutiply-adds;
\item
	With lookup tables: $256\times d$ multiply-adds and $n\times b$ lookup-adds.
\end{itemize}
This is the key to the efficiency of the product quantizer.
In our GPU implementation, $b$ is any multiple of 4 up to 64. The codes are stored as sequential groups of $b$ bytes per vector within lists.

\newcommand{\listno}{L}

\paragraph{IVFADC lookup tables.}
\label{sec:threeterms}
When scanning over the elements of the inverted list $\invlist_\listno$ (where by definition $q_1(y)$ is constant), the look-up table method can be applied, as the query $x$ and $q_1(y)$ are known.

Moreover, the computation of the tables $T_1\dots T_b$ is further optimized~\cite{BL14a}.
The expression of $\|x - q(y)\|_2^2$ in Equation (\ref{eq:twostagedecomp}) can be decomposed as:
\begin{equation}
\underbrace{\|q_2(...)\|_2^2 + 2\langle q_1(y),q_2(...)\rangle}_{\textrm{term 1}}
+
\underbrace{\|x - q_1(y)\|_2^2}_{\textrm{term 2}}
-2
\underbrace{\langle x,q_2(...)\rangle}_{\textrm{term 3}}.
\label{eq:precomputed}
\end{equation}

The objective is to minimize inner loop computations. The computations we can do in advance and store in lookup tables are as follows:
\begin{itemize}
\item
	Term 1 is independent of the query. It can be precomputed from the quantizers, and stored in a table $\mathcal{T}$ of size $|\mathcal{C}_1|\times 256 \times b$;
\item
	Term 2 is the distance to $q_1$'s reproduction value. It is thus a by-product of the first-level quantizer $q_1$;
\item
	Term 3 can be computed independently of the inverted list. Its computation costs $d\times 256$ multiply-adds.
\end{itemize}
This decomposition is used to produce the lookup tables $T_1 \dots T_b$ used during the scan of the inverted list.
For a single query, computing the $\nprobe \times b$ tables from scratch costs $\nprobe \times d \times 256$ multiply-adds, while this decomposition costs $256\times d$ multiply-adds and $\nprobe \times b \times 256$ additions. On the GPU, the memory usage of $\mathcal{T}$ can be prohibitive, so we enable the decomposition only when memory is a not a concern.

\subsection{GPU implementation}
Algorithm~\ref{alg:ivfpq} summarizes the process as one would implement it on a CPU. The inverted lists are stored as two separate arrays, for PQ codes and associated IDs. IDs are resolved only if $k$-selection determines $k$-nearest membership. This lookup yields a few sparse memory reads in a large array, thus the IDs can optionally be stored on CPU for tiny performance cost.

\paragraph{List scanning.}
A kernel is responsible for scanning the $\nprobe$ closest inverted lists for each query, and calculating the per-vector pair distances using the lookup tables $T_i$. The $T_i$ are stored in shared memory: up to $n_q \times \nprobe \times \max_i|\invlist_i| \times b$ lookups are required for a query set (trillions of accesses in practice), and are random access. This limits $b$ to at most 48 (32-bit floating point) or 96 (16-bit floating point) with current architectures. In case we do not use the decomposition of Equation~(\ref{eq:precomputed}), the $T_i$ are calculated by a separate kernel before scanning.

\paragraph{Multi-pass kernels.}
Each $n_q \times \nprobe$ pairs of query against inverted list can be processed independently. At one extreme, a block is dedicated to each of these, resulting in up to $n_q \times \nprobe \times \max_i|\invlist_i|$ partial results being written back to global memory, which is then $k$-selected to $n_q \times k$ final results. This yields high parallelism but can exceed available GPU global memory; as with exact search, we choose a tile size $t_q \leq n_q$ to reduce memory consumption, bounding its complexity by $\mathcal{O}(2t_q \nprobe \max_i |\invlist_i| )$ with multi-streaming.

A single warp could be dedicated to $k$-selection of each $t_q$ set of lists, which could result in low parallelism. We introduce a two-pass $k$-selection, reducing $t_q \times \nprobe \times \max_i|\invlist_i|$ to $t_q \times f \times k$ partial results for some subdivision factor $f$. This is reduced again via $k$-selection to the final $t_q \times k$ results.

\paragraph{Fused kernel.}
As with exact search, we experimented with a kernel that dedicates a single block to scanning all $\nprobe$ lists for a single query, with $k$-selection fused with distance computation. This is possible as \textsc{WarpSelect} does not fight for the shared memory resource which is severely limited. This reduces global memory write-back, since almost all intermediate results can be eliminated. However, unlike $k$-selection overhead for exact computation, a significant portion of the runtime is the gather from the $T_i$ in shared memory and linear scanning of the $\invlist_i$ from global memory; the write-back is not a dominant contributor. Timing for the fused kernel is improved by at most 15\%, and for some problem sizes would be subject to lower parallelism and worse performance without subsequent decomposition. Therefore, and for reasons of implementation simplicity, we do not use this layout.

\begin{algorithm}
  \caption{IVFPQ batch search routine}\label{alg:ivfpq}
  \begin{algorithmic}
    \Function{ivfpq-search}{$[x_1, ... , x_\nq]$, $\invlist_1,...,\invlist_{|\mathcal{C}_1|}$}

	\For{$i \gets 0 : \nq$} \Comment batch quantization of Section~\ref{sec:exactimpl}
	\State $L_\mathrm{IVF}^i \gets \nprobe\text{-}\textrm{argmin}_{c \in \mathcal{C}_1} \| x - c \|_2$
	\EndFor
	\For{$i \gets 0 : \nq$}
	\State $L \gets []$ \Comment distance table
	\State Compute term 3 (see Section~\ref{sec:threeterms})
	       \For{$\listno$ in $L_\mathrm{IVF}^i$} \Comment $\nprobe$ loops
	       \State Compute distance tables $T_1,..., T_b$
	              \For{$j$ in $\invlist_\listno$}
		          \State \Comment distance estimation, Equation (\ref{eq:pqsum})
	              \State  $d\gets \|x_i - q(y_j)\|^2_2$
	              \State  Append $(d, \listno, j)$ to $L$
 	              \EndFor
	       \EndFor
	\State $R_i\gets$ k-select smallest distances $d$ from $L$
	\EndFor
\State\Return R
    \EndFunction
  \end{algorithmic}
\end{algorithm}

\subsection{Multi-GPU parallelism}
Modern servers can support several GPUs. We employ this capability for both compute power and memory.

\paragraph{Replication.} If an index instance fits in the memory of a single GPU, it can be replicated across $\mathcal{R}$ different GPUs. To query $\nq$ vectors, each replica handles a fraction $\nq / \mathcal{R}$ of the queries, joining the results back together on a single GPU or in CPU memory. Replication has near linear speedup, except for a potential loss in efficiency for small $\nq$.

\paragraph{Sharding.} If an index instance does not fit in the memory of a single GPU, an index can be sharded across $\mathcal{S}$ different GPUs. For adding $\ell$ vectors, each shard receives $\ell / \mathcal{S}$ of the vectors, and for query, each shard handles the full query set $\nq$, joining the partial results (an additional round of $k$-selection is still required) on a single GPU or in CPU memory. For a given index size $\ell$, sharding will yield a speedup (sharding has a query of $\nq$ against $\ell / \mathcal{S}$ versus replication with a query of $\nq / \mathcal{R}$ against $\ell$), but is usually less than pure replication due to fixed overhead and cost of subsequent $k$-selection.

Replication and sharding can be used together ($\mathcal{S}$ shards, each with $\mathcal{R}$ replicas for $\mathcal{S} \times \mathcal{R}$ GPUs in total). Sharding or replication are both fairly trivial, and the same principle can be used to distribute an index across multiple machines.

\newpage
\section{Experiments \& Applications}
\label{sec:experiments}

This section compares our GPU $k$-selection and nearest-neighbor approach to existing libraries. Unless stated otherwise, experiments are carried out on a 2$\times$2.8GHz Intel Xeon E5-2680v2 with 4 Maxwell Titan X GPUs on CUDA 8.0.

\subsection{k-selection performance}
\label{sec:kperf}

We compare against two other GPU small $k$-selection implementations: the row-based Merge Queue with Buffered Search and Hierarchical Partition extracted from the \textit{fgknn} library of Tang et al.~\cite{Tang15} and Truncated Bitonic Sort (\textit{TBiS}) from Sismanis et al.~\cite{Sismanis12}. Both were extracted from their respective exact search libraries.

We evaluate $k$-selection for $k = 100$ and 1000 of each row from a row-major matrix $n_q \times \ell$ of random  32-bit floating point values on a single Titan X. The batch size $n_q$ is fixed at 10000, and the array lengths $\ell$ vary from 1000 to 128000. Inputs and outputs to the problem remain resident in GPU memory, with the output being of size $n_q \times k$, with corresponding indices. Thus, the input problem sizes range from 40\,MB ($\ell$\,=\,$1000$) to 5.12\,GB ($\ell$\,=\,$128$k). TBiS requires large auxiliary storage, and is limited to $\ell \leq 48000$ in our tests.

%
%
%
%
%
%

\begin{figure}
\includegraphics[width=\linewidth]{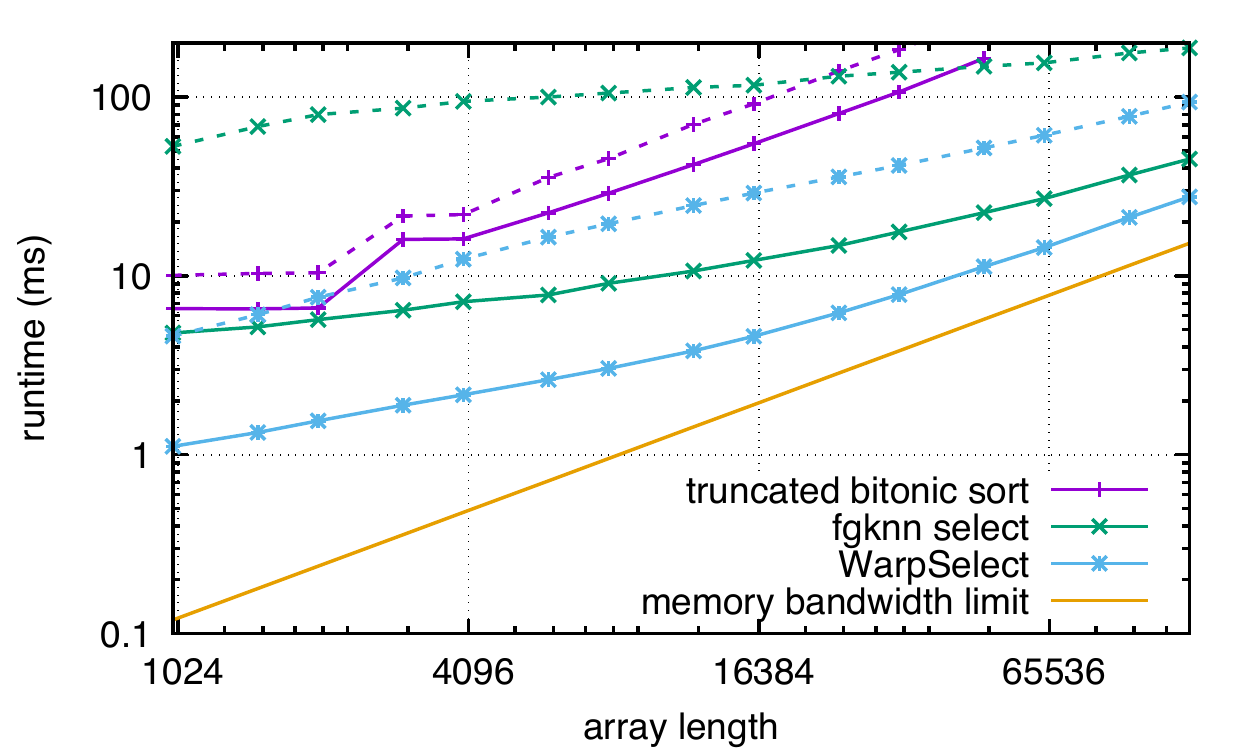}
\caption{\label{fig:raw100x}
	Runtimes for different $k$-selection methods, as a function of array length $\ell$. Simultaneous arrays processed are $n_q = 10000$. $k=100$ for full lines, $k=1000$ for dashed lines.
}
\end{figure}

Figure~\ref{fig:raw100x} shows our relative performance against TBiS and fgknn. It also includes the peak possible performance given by the memory bandwidth limit of the Titan X. The relative performance of \warpbased over fgknn increases for larger $k$; even TBiS starts to outperform fgknn for larger $\ell$ at $k = 1000$. We look especially at the largest $\ell = 128000$. \warpbased is $1.62 \times$ faster at $k = 100$, $2.01 \times$ at $k = 1000$. Performance against peak possible drops off for all implementations at larger $k$. \warpbased operates at \textbf{55\% of peak} at $k = 100$ but only 16\% of peak at $k = 1000$. This is due to additional overhead assocated with bigger thread queues and merge/sort networks for large $k$.

\paragraph{Differences from fgknn.}
\warpbased is influenced by fgknn, but has several improvements: all state is maintained in registers (no shared memory), no inter-warp synchronization or buffering is used, no ``hierarchical partition'', the $k$-selection can be fused into other kernels, and it uses odd-size networks for efficient merging and sorting.
\bigskip

\subsection{k-means clustering}

The exact search method with $k=1$ can be used by a $k$-means clustering method in the assignment stage, to assign $\nq$ training vectors to $|\mathcal{C}_1|$ centroids. Despite the fact that it does not use the IVFADC and $k=1$ selection is trivial (a parallel reduction is used for the $k = 1$ case, not \warpbased), $k$-means is a good benchmark for the clustering used to train the quantizer $q_1$.

We apply the algorithm on MNIST8m images. The 8.1M images are graylevel digits in 28x28 pixels, linearized to vectors of 784-d. We compare this $k$-means implementation to the GPU $k$-means of BIDMach~\cite{CZ13}, which was shown to be more efficient than several distributed $k$-means implementations that require dozens of machines\footnote{BIDMach numbers from \url{https://github.com/BIDData/BIDMach/wiki/Benchmarks\#KMeans}}.
Both algorithms were run for 20 iterations. Table~\ref{tab:kmeans} shows that our implementation is more than \textbf{2$\times$ faster}, although both are built upon cuBLAS. Our implementation receives some benefit from the $k$-selection fusion into L2 distance computation. For multi-GPU execution via replicas, the speedup is close to linear for large enough problems (3.16$\times$ for 4 GPUs with 4096 centroids).
Note that this benchmark is somewhat unrealistic, as one would typically sub-sample the dataset randomly when so few centroids are requested.

\begin{table}
{
\centering \begin{tabular}{lcrr}
\hline
       &  & \multicolumn{2}{c}{\# centroids} \\
method & \# GPUs      & 256\, & 4096\, \\
\hline
BIDMach~\cite{CZ13}& 1 & 320\,s  & 735\,s \\
Ours   & 1 & 140\,s & 316\,s  \\
Ours   & 4 & 84\,s  & 100\,s  \\
\hline
\end{tabular}
\caption{\label{tab:kmeans}
MNIST8m $k$-means performance
}}
\end{table}

\paragraph{Large scale.}
We can also compare to~\cite{AKAE15}, an approximate CPU method that clusters $10^8$ 128-d vectors to 85k centroids. Their clustering method runs in 46~minutes, but requires 56~minutes (at least) of pre-processing to encode the vectors. Our method performs \emph{exact} k-means on 4~GPUs in 52~minutes without any pre-processing.

\subsection{Exact nearest neighbor search}

\begin{figure}[t]
\includegraphics[width=\linewidth]{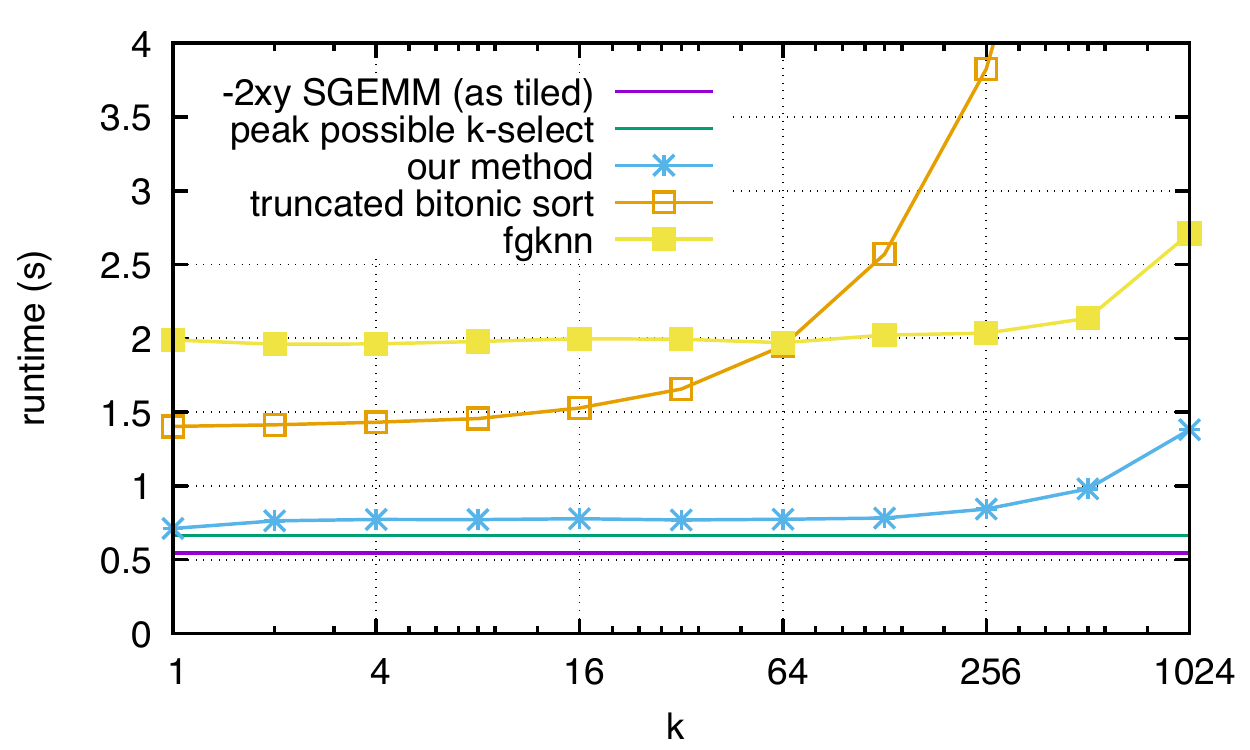}
\caption{\label{fig:cmpkselect}
  Exact search $k$-NN time for the SIFT1M dataset with varying $k$ on 1 Titan X GPU. }
\end{figure}

We consider a classical dataset used to evaluate nearest neighbor search: \textsc{Sift1M}~\cite{JDS11}. Its characteristic sizes are $\ell=10^6$, $d=128$, $\nq=10^4$. Computing the partial distance matrix $D'$ costs $\nq \times \ell \times d = 1.28$ Tflop, which runs in less than one second on current GPUs. Figure~\ref{fig:cmpkselect} shows the cost of the distance computations against the cost of our tiling of the GEMM for the $ - 2 \left< x_j, y_i \right>$ term of Equation~\ref{eq:dminmmul} and the peak possible $k$-selection performance on the distance matrix of size $\nq \times \ell$, which additionally accounts for reading the tiled result matrix $D'$ at peak  memory bandwidth.

In addition to our method from Section~\ref{sec:layout}, we include times from the two GPU libraries evaluated for $k$-selection performance in Section~\ref{sec:kperf}.
We make several observations:
\begin{itemize}
\item
	for $k$-selection, the naive algorithm that sorts the full result array for each query using \verb|thrust::sort_by_key| is more than $10\times$ slower than the comparison methods;
\item
  L2 distance and $k$-selection cost is dominant for all but our method, which has \textbf{85~\% of the peak} possible performance, assuming GEMM usage and our tiling of the partial distance matrix $D'$ on top of GEMM is close to optimal. The cuBLAS GEMM itself has low efficiency for small reduction sizes ($d = 128$);
\item
  Our fused L2/$k$-selection kernel is important. Our same exact algorithm without fusion (requiring an additional pass through $D'$) is at least 25\% slower.

\end{itemize}

Efficient $k$-selection is even more important in situations where approximate methods are used to compute distances, because the relative cost of $k$-selection with respect to distance computation increases. %

\subsection{Billion-scale approximate search}
\label{sec:largescale}

There are few studies on GPU-based approximate nearest-neighbor search on large datasets ($\ell \gg 10^6$). We report a few comparison points here on index search, using standard datasets and evaluation protocol in this field.

\paragraph{SIFT1M.} For the sake of completeness, we first compare our GPU search speed on \textsc{Sift1M} with the implementation of Wieschollek et al.~\cite{WieschollekCVPR16}. They obtain a nearest neighbor recall at 1 (fraction of queries where the true nearest neighbor is in the top 1 result) of R@1\,=\,0.51, and R@100\,=\,0.86 in 0.02\,ms per query on a Titan X. For the same time budget, our implementation obtains R@1\,=\,0.80 and R@100\,=\,0.95.

\paragraph{SIFT1B.}
We compare again with Wieschollek et al., on the \textsc{Sift1B} dataset~\cite{JTDA11} of 1 billion SIFT image features at $\nq = 10^4$. We compare the search performance in terms of same memory usage for similar accuracy (more accurate methods may involve greater search time or memory usage). On a single GPU, with $m = 8$ bytes per vector, R@10\,=\,0.376 in 17.7\,$\mu$s per query vector, versus their reported R@10\,=\,0.35 in 150\,$\mu$s per query vector. Thus, our implementation is more accurate at a speed \textbf{8.5$\times$ faster}.

\paragraph{DEEP1B.}
We also experimented on the \textsc{Deep1B} dataset~\cite{Babenko16} of $\ell$=1 billion CNN representations for images at $\nq = 10^4$. The paper that introduces the dataset reports CPU results (1 thread): R@1\,=\,0.45 in 20\,ms search time per vector. We use a PQ encoding of $m = 20$, with $d = 80$ via OPQ~\cite{GHKS14}, and $|\mathcal{C}_1|=2^{18}$, which uses a comparable dataset storage as the original paper (20 GB). This requires multiple GPUs as it is too large for a single GPU's global memory, so we consider 4 GPUs with $\mathcal{S} = 2$, $\mathcal{R} = 2$. We obtain a R@1\,=\,0.4517 in 0.0133\,ms per vector. While the hardware platforms are different, it shows that making searches on GPUs is a game-changer in terms of speed achievable on a single machine.

\begin{figure}
\includegraphics[width=\linewidth]{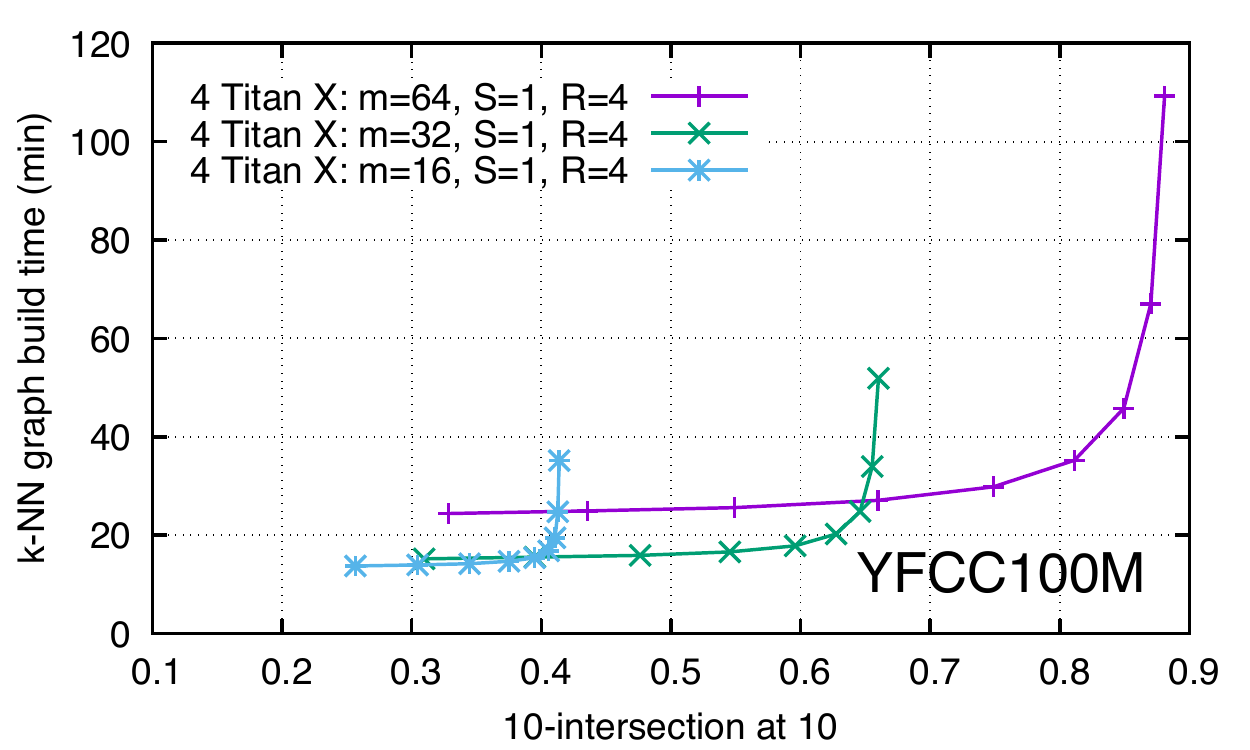}
\includegraphics[width=\linewidth]{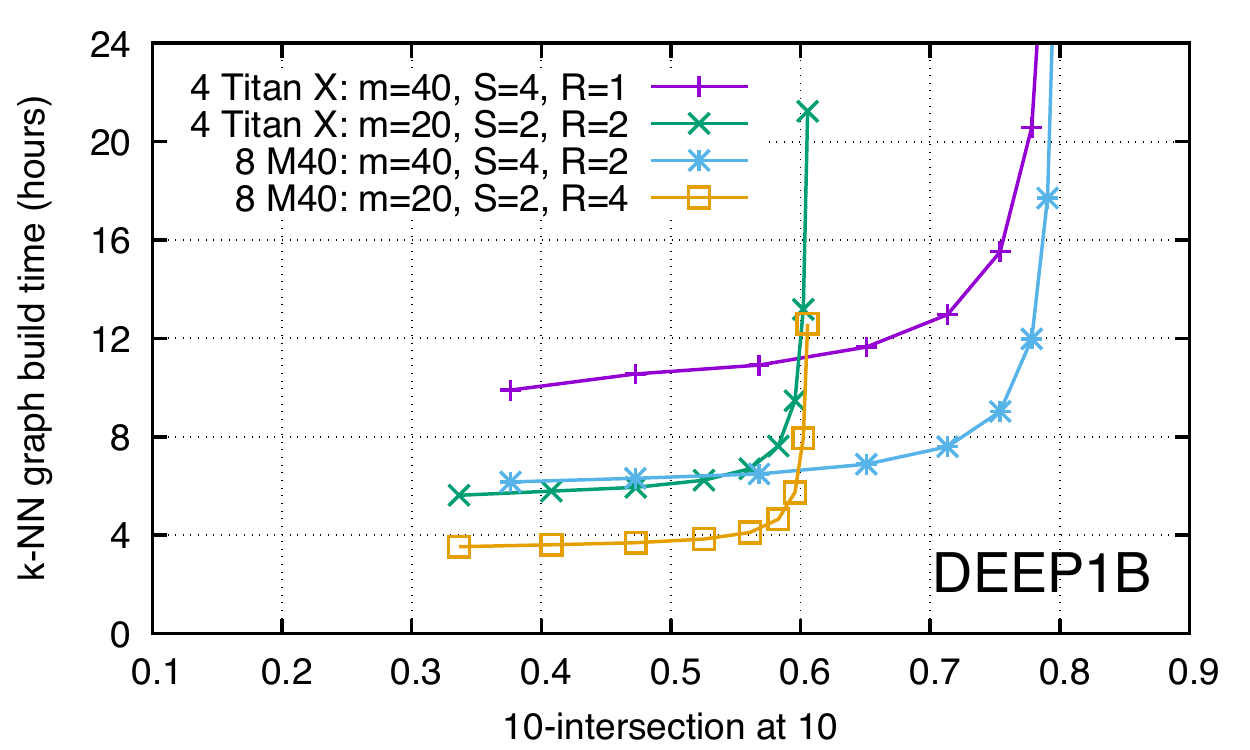}
\caption{\label{fig:knngraph}
        Speed/accuracy trade-off of brute-force 10-NN graph construction for the YFCC100M and DEEP1B datasets.
}
\end{figure}

\begin{figure*}
\includegraphics[width=\linewidth]{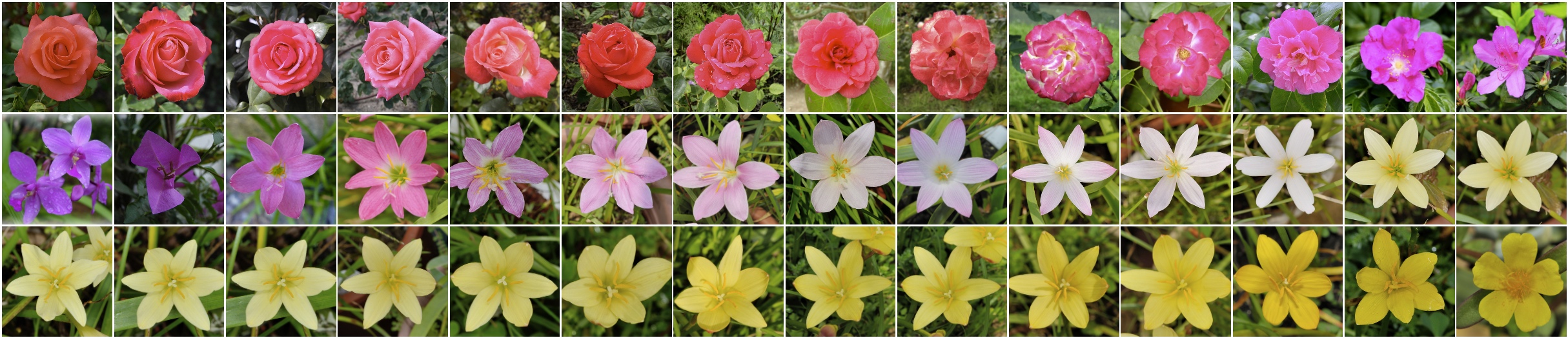}
\vspace{-10pt}
\caption{\label{fig:path}
	Path in the $k$-NN graph of 95 million images from YFCC100M. The first and the last image are given; the algorithm computes the smoothest path between them.
}
\end{figure*}

\subsection{The k-NN graph}

An example usage of our similarity search method is to construct a \textit{$k$-nearest neighbor graph} of a dataset via brute force (all vectors queried against the entire index).

\paragraph{Experimental setup.}
We evaluate the trade-off between speed, precision and memory  on two datasets: 95 million images from the \textsc{Yfcc100M} dataset~\cite{Thomee16} and \textsc{Deep1B}. For \textsc{Yfcc100M}, we compute CNN descriptors as the one-before-last layer of a ResNet~\cite{He16}, reduced to $d$\,=\,128 with PCA.

The evaluation measures the trade-off between:
\begin{itemize}
\item
  Speed: How much time it takes to build the IVFADC index from scratch and construct the whole $k$-NN graph ($k = 10$) by searching nearest neighbors for all vectors in the dataset. Thus, this is an end-to-end test that includes indexing as well as search time;
\item
  Quality: We sample 10,000 images for which we compute the exact nearest neighbors. Our accuracy measure is the fraction of 10 found nearest neighbors that are within the ground-truth 10 nearest neighbors.
\end{itemize}

\noindent
For \textsc{Yfcc100M}, we use a coarse quantizer ($2^{16}$ centroids), and consider $m = $ 16, 32 and 64 byte PQ encodings for each vector. For \textsc{Deep1B}, we pre-process the vectors to $d=120$ via OPQ, use $|\mathcal{C}_1| = 2^{18}$ and consider $m = $ 20, 40. For a given encoding, we vary $\nprobe$ from 1 to 256, to obtain trade-offs between efficiency and quality, as seen in Figure~\ref{fig:knngraph}.

\paragraph{Discussion.}
For \textsc{Yfcc100M} we used $\mathcal{S} = 1$, $\mathcal{R} = 4$. An accuracy of more than 0.8 is obtained in 35~minutes.
For \textsc{Deep1B}, a lower-quality graph can be built in 6 hours, with higher quality in about half a day. We also experimented with more GPUs by doubling the replica set, using 8 Maxwell M40s (the M40 is roughly equivalent in performance to the Titan X). Performance is improved sub-linearly ($\sim1.6\times$ for $m=20$, $\sim1.7\times$ for $m=40$).

For comparison, the largest $k$-NN graph construction we are aware of used a dataset comprising 36.5 million 384-d vectors, which took a cluster of 128 CPU servers 108.7 hours of compute~\cite{Warashina14}, using NN-Descent~\cite{DCL11}. Note that NN-Descent could also build or refine the $k$-NN graph for the datasets we consider, but it has a large memory overhead over the graph storage, which is already 80~GB for \textsc{Deep1B}. Moreover it requires random access across all vectors (384\,GB for \textsc{Deep1B}).

The largest GPU $k$-NN graph construction we found is a brute-force construction using exact search with GEMM, of a dataset of 20 million 15,000-d vectors, which took a cluster of 32 Tesla C2050 GPUs 10 days~\cite{Dashti13}. Assuming computation scales with GEMM cost for the distance matrix, this approach for \textsc{Deep1B} would take an impractical 200 days of computation time on their cluster.

\subsection{Using the k-NN graph}

When a $k$-NN graph has been constructed for an image dataset, we can find paths in the graph between any two images, provided there is a single connected component (this is the case). For example, we can search the shortest path between two images of flowers, by propagating neighbors from a starting image to a destination image. Denoting by $S$ and $D$ the source and destination images, and $d_{ij}$ the distance between nodes, we search the path $P=\{p_1, ..., p_n\}$ with $p_1=S$ and $p_n=D$ such that
\begin{equation}
\min_{P}\max_{i=1..n} d_{p_{i}p_{i+1}},
\end{equation}
\textit{i.e.}, we want to favor smooth transitions. An example result is shown in Figure~\ref{fig:path} from \textsc{Yfcc100M}\footnote{The mapping from vectors to images is not available for \textsc{Deep1B}}. It was obtained after 20~seconds of propagation in a $k$-NN graph with $k=15$ neighbors. Since there are many flower images in the dataset, the transitions are smooth.
\pagebreak

\section{Conclusion}

The arithmetic throughput and memory bandwidth of GPUs are well into the teraflops and hundreds of gigabytes per second.
However, implementing algorithms that approach these performance levels is complex and counter-intuitive.
In this paper, we presented the algorithmic structure of similarity search methods that achieves near-optimal performance on GPUs.

This work enables applications that needed complex approximate algorithms before. For example, the approaches presented here make it possible to do exact $k$-means clustering or to compute the $k$-NN graph with simple brute-force approaches in less time than a CPU (or a cluster of them) would take to do this approximately.

GPU hardware is now very common on scientific workstations, due to their popularity for machine learning algorithms.
We believe that our work further demonstrates their interest for database applications.
Along with this work, we are publishing a carefully engineered implementation of this paper's algorithms, so that these GPUs can now also be used for efficient similarity search.

\balance

{\small
\bibliographystyle{abbrv}
\bibliography{egbib}
}

\section*{Appendix: Complexity analysis of \warpbased}

We derive the average number of times updates are triggered in \warpbased, for use in Section~\ref{sec:k-sel-complexity}.

Let the input to $k$-selection be a sequence $\{a_1, a_2, ..., a_{\ell}\}$ (1-based indexing), a randomly chosen permutation of a set of distinct elements.
Elements are read sequentially in $c$ groups of size $w$ (the warp; in our case, $w = 32$); assume $\ell$ is a multiple of $w$, so $c = \ell / w$. Recall that $t$ is the thread queue length.
We call elements prior to or at position $n$ in the min-$k$ seen so far the \textit{successive min-$k$ (at $n$)}.
The likelihood that $a_n$ is in the successive min-$k$ at $n$ is:

\begin{equation} \alpha(n, k) :=
  \begin{cases}
    1       & \quad \text{if } n \leq k \\
    k/n     & \quad \text{if } n > k \\
  \end{cases}
\end{equation}
as each $a_n$, $n > k$ has a $k/n$ chance as all permutations are equally likely, and all elements in the first $k$ qualify.

\paragraph{Counting the insertion sorts.}
In a given lane, an insertion sort is triggered if the incoming value is in the successive min-$k + t$ values, but the lane has ``seen'' only $wc_0 + (c-c_0)$ values, where $c_0$ is the previous won warp ballot. The probability of this happening is:
\begin{equation}
\alpha(wc_0 + (c-c_0), k + t)  \approx  %
\frac{k + t}{wc} \textrm{  for  } c > k.
\end{equation}
The approximation considers that the thread queue has seen \emph{all} the $wc$ values, not just those assigned to its lane. The probability of \emph{any} lane triggering an insertion sort is then:
\begin{equation}
1 - \left( 1 - \frac{k + t}{wc}\right)^w \approx \frac{k + t}{c}.
\end{equation}
Here the approximation is a first-order Taylor expansion.
Summing up the probabilities over $c$ gives an expected number of insertions of $N_2 \approx (k+t)\log(c)=\mathcal{O}(k\log(\ell/w))$.

\paragraph{Counting full sorts.}
 We seek $N_3 =\pi(\ell, k, t, w)$, the expected number of full sorts required for \warpbased. %

\paragraph{Single lane.}
For now, we assume $w = 1$, so $c = \ell$. Let $\gamma(\ell, m, k)$ be the probability that in an sequence $\{a_1, ..., a_\ell\}$, exactly $m$ of the elements as encountered by a sequential scanner ($w = 1$) are in the successive min-$k$. Given $m$, there are $\ell \choose m$ places where these successive min-$k$ elements can occur. It is given by a recurrence relation:

\begin{equation} \gamma(\ell, m, k) :=
\begin{cases}
  1       & \hspace{-11pt} \ell = 0 \text { and } m = 0\\
  0       & \hspace{-11pt} \ell = 0 \text { and } m > 0\\
  0       & \hspace{-11pt} \ell > 0 \text { and } m = 0\\
  (\gamma(\ell - 1, m - 1, k) \cdot \alpha(\ell, k) + \\\gamma(\ell - 1, m, k) \cdot (1 - \alpha(\ell, k))) & \text{otherwise}. \\
  \end{cases}
\end{equation}

The last case is the probability of: there is a $\ell-1$ sequence with $m-1$ successive min-$k$ elements preceding us, and the current element is in the successive min-$k$, or the current element is not in the successive min-$k$, $m$ ones are before us. We can then develop a recurrence relationship for $\pi(\ell, k, t, 1)$. Note that

\begin{equation}
\delta(\ell, b, k, t) :=
\sum_{m = bt}^{\min((bt + \max(0, t - 1)), \ell)} \gamma(\ell, m, k)
\end{equation}\\
for $b$ where $0 \leq bt \leq \ell$ is the fraction of all sequences of length $\ell$ that will force $b$ sorts of data by winning the thread queue ballot, as there have to be $bt$ to $(bt + \max(0, t - 1))$ elements in the successive min-$k$ for these sorts to happen (as the min-$k$ elements will overflow the thread queues). There are at most $\left \lfloor \ell / t \right \rfloor$ won ballots that can occur, as it takes $t$ separate sequential current min-$k$ seen elements to win the ballot. $\pi(\ell, k, t, 1)$ is thus the expectation of this over all possible $b$:

\begin{equation}
\pi(\ell, k, t, 1) = \sum_{b=1}^{\left \lfloor \ell / t \right \rfloor} b \cdot \delta(\ell, b, k, t).
\end{equation}
This can be computed by dynamic programming.
Analytically, note that for $t = 1$, $k = 1$, $\pi(\ell, 1, 1, 1)$ is the harmonic number $H_\ell = 1 + \frac{1}{2} + \frac{1}{3} + ... + \frac{1}{\ell}$, which converges to $\ln(\ell) + \gamma$ (the Euler-Mascheroni constant $\gamma$) as $\ell \rightarrow \infty$.\\

For $t = 1, k > 1, \ell > k$, $\pi(\ell, k, 1, 1) = k + k (H_{\ell} - H_k)$ or $\mathcal{O}(k \log(\ell))$, as the first $k$ elements are in the successive min-$k$, and the expectation for the rest is $\frac{k}{k+1} + \frac{k}{k+2} + ... + \frac{k}{\ell}$.\\

For $t > 1, k > 1, \ell > k$, note that there are some number $D$, $k \leq D \leq \ell$ of successive min-$k$ determinations $D$ made for each possible $\{a_1, ..., a_\ell\}$. The number of won ballots for each case is by definition $\left \lfloor D / t \right \rfloor$, as the thread queue must fill up $t$ times. Thus, $\pi(\ell, k, t, 1) = \mathcal{O}(k \log(\ell) / t)$.

\paragraph{Multiple lanes.}
The $w > 1$ case is complicated by the fact that there are joint probabilities to consider (if more than one of the $w$ workers triggers a sort for a given group, only one sort takes place). %
However, the likelihood can be bounded.
Let $\pi'(\ell, k, t, w)$ be the expected won ballots assuming no mutual interference between the $w$ workers for winning ballots (i.e., we win $b$ ballots if there are $b \leq w$ workers that independently win a ballot at a single step), but with the shared min-$k$ set after each sort from the joint sequence.
Assume that $k \geq w$. Then:

\begin{equation}
\begin{aligned}
\pi'(\ell, k, 1, w) & \leq w \Bigg( \left \lceil \frac{k}{w} \right \rceil + \sum_{i=1}^{\left \lceil \ell / w \right \rceil - \left \lceil k/w \right \rceil} \frac{k}{w(\left \lceil k / w \right \rceil + i)} \Bigg)\\
& \leq w \pi(\left \lceil \ell / w \right \rceil\hspace{-2pt}, k, 1, 1) = \mathcal{O}(wk \log(\ell / w))
\end{aligned}
\end{equation}
where the likelihood of the $w$ workers seeing a successive min-$k$ element has an upper bound of that of the first worker at each step. As before, the number of won ballots is scaled by $t$, so $\pi'(\ell, k, t, w) = \mathcal{O}(wk \log(\ell / w) / t)$. Mutual interference can only reduce the number of ballots, so we obtain the same upper bound for $\pi(\ell, k, t, w)$.

\end{document}